\documentclass[a4paper,twoside]{article}
\usepackage{epsfig}
\usepackage{subcaption}
\usepackage{calc}
\usepackage{amssymb}
\usepackage{amstext}
\usepackage{amsmath}
\usepackage{amsthm}
\usepackage{multicol}
\usepackage{pslatex}
\usepackage[sort]{natbib}
\usepackage{algorithm2e}
\usepackage[bottom]{footmisc}
\usepackage{graphicx}
\usepackage{multirow}
\usepackage[dvipsnames]{xcolor}
\usepackage{pifont}
\usepackage{url}
\usepackage{SCITEPRESS}     

\begin{document}

\title{EigDiff: Eigensample Diffusion -- or how to significantly reduce training dataset size with marginal performance variation?}
\author{\authorname{Andrei Jarca, Vlad Hondru, Florinel-Alin Croitoru, Radu Tudor Ionescu}
\affiliation{University of Bucharest, Romania}
\email{Corresponding author: raducu.ionescu@gmail.com}
}


\abstract{Diffusion models are notoriously expensive to train, requiring large-scale datasets to learn how to gradually reverse the noise from images over multiple small denoising steps. Current approaches focus on distilling the number of diffusion steps, neglecting potential redundancies of the training data. In this paper, we explore the potential of harnessing eigensamples to train diffusion models on significantly smaller datasets, while preserving model performance. To mitigate the sensitivity to affine transformations of eigenimages, we harness the latent space of a pre-trained CLIP model. We next compute latent eigenvectors by performing principal component analysis (PCA) in CLIP space. Further, we employ eigenvectors as reference points to guide the selection of a representative training set comprising comprising both exemplars (images whose embeddings are closest to eigenvectors) and diverse images. Our comprehensive experiments with different diffusion models applied on multiple datasets indicate that our approach (EigDiff) can reduce training data by up to $14\times$ and accelerate subset selection by up to $1607\times$, with marginal performance differences. 
\vspace{-0.25cm}}

\onecolumn \maketitle \normalsize \setcounter{footnote}{0} \vfill

\setlength{\abovedisplayskip}{2.0pt}
\setlength{\belowdisplayskip}{2.0pt}
\setlength{\abovedisplayshortskip}{2.0pt}
\setlength{\belowdisplayshortskip}{2.0pt}

\vspace{-0.2cm}
\section{\uppercase{Introduction}}
\label{sec:introduction}
\vspace{-0.2cm}

Diffusion models~\citep{Croitoru-TPAMI-2023,song-NeurIPS-2019,song-ICLR-2021,dhariwal-NeurIPS-2021,ho-NeurIPS-2020,rombach-CVPR-2022,Saharia-NeurIPS-2022,song-ICLR-2021b,Peebles-ICCV-2023} have consistently demonstrated state-of-the-art performance in text-to-image synthesis in recent years. Some of the recent developments of diffusion models have focused on improving the quality of the generated samples. This is often achieved by a specialized post-training or fine-tuning phase~\citep{Zhang-ECCV-2024,Luo-NeurIPS-2025} designed to improve the base models toward better visual and semantic results. This refinement process often involves continued training~\citep{dai-arXiv-2023} on highly curated and/or aesthetically superior datasets to eliminate structural artifacts and enhance fine-grained details. Furthermore, to bridge the gap between raw generative capability and complex user intent, researchers integrate human preference alignment and reinforcement learning methods \citep{Croitoru-CVPR-2025,Wallace-arxiv-2023,na-CVPR-2025,Black-ICLR-2024,Fan-NeurIPS-2023,Liu-NeurIPS-2026,Ren-ICLR-2025}. By directly optimizing for human-based reward signals, these advanced alignment strategies enhance structural coherence, aesthetic appeal, and strict prompt fidelity. Usually, the fine-tuning process operates on a massive data scale of its own. Although foundational models have already undergone rigorous pre-training on web-scale datasets, specialized datasets utilized during the fine-tuning phase remain exceptionally large \citep{Kirstain-NeurIPS-2023,Wu-arXiv-2023}. In this work, we show that the size of these datasets can be substantially reduced through a representative selection of samples from the original set.

\begin{figure*}[!t]
    \centering
    \includegraphics[width=0.9\linewidth]{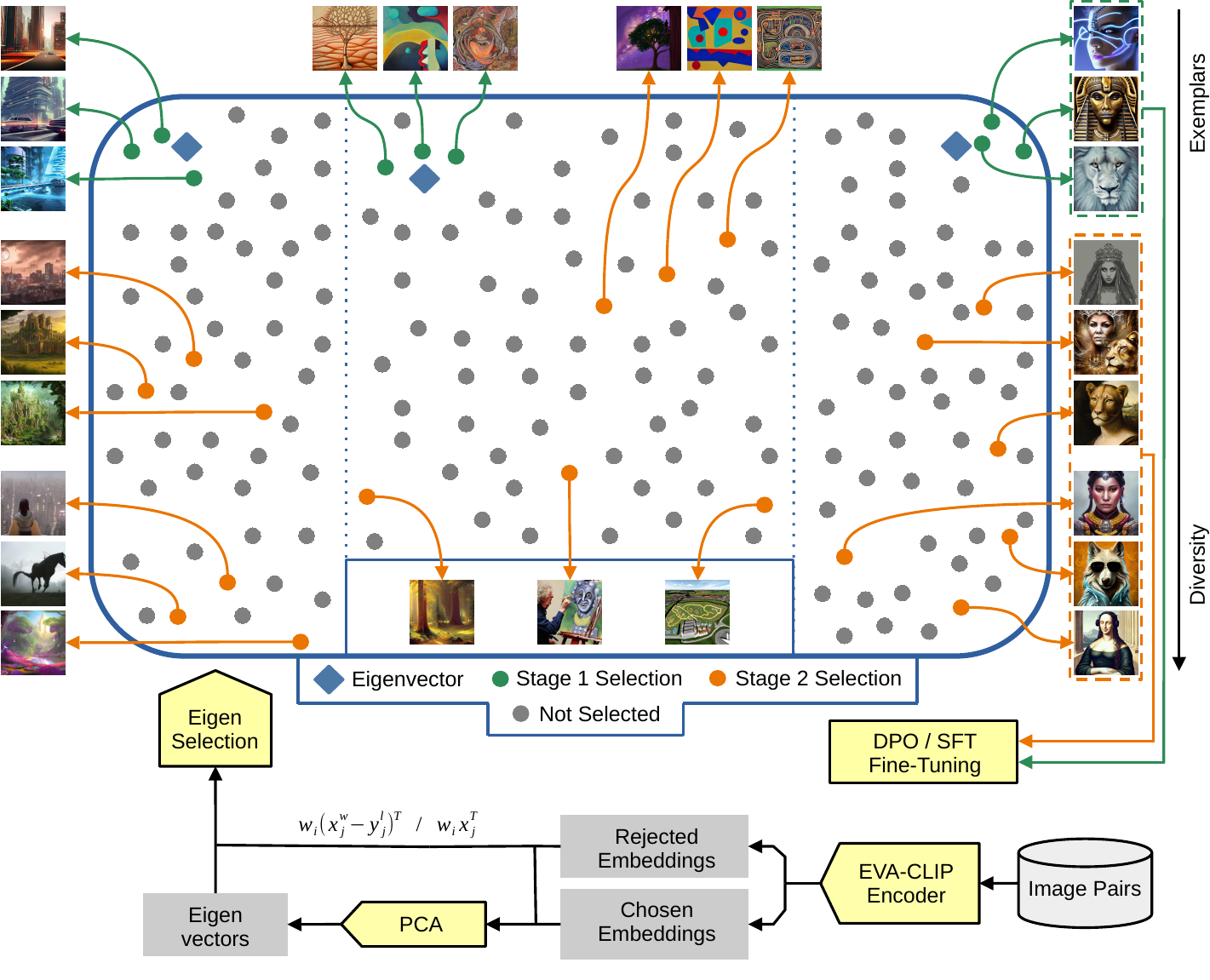}
    \vspace{-0.25cm}
    \caption{Overview of EigDiff (eigen-guided data selection). Preference image pairs are CLIP-encoded into embeddings, on which PCA yields the principal (eigen) components. Each pair is scored by its projection onto these directions ($\mathbf{w}\cdot\mathbf{x}^\top$ for SFT, $\mathbf{w}\cdot(\mathbf{x}-\mathbf{y})^\top$ for DPO). Per component, Stage~1 deterministically selects the exemplars (highest score). The figure illustrates the near to far spread. Stage~2 stratifies the full remaining score range via diversity quantiles. Best viewed in color.
}
\vspace{-0.25cm}
    \label{fig:placeholder123}
\end{figure*}

Despite the remarkable generative capabilities of diffusion models, the inherent latency of their iterative reverse process represents a significant drawback. Consequently, another important research direction has been the aggressive optimization of the inference phase, targeting a drastic reduction in the number of denoising steps required to generate an image \citep{Song-ICLR-2024,Song-ICML-2023,Luo-arXiv-2023,salimans-arXiv-2022,Yin-NeurIPS-2024,Yin-CVPR-2024,Sauer-ECCV-2024,Frans-ICLR-2025}. The most popular methods to achieve this kind of accelerations are based on knowledge distillation \citep{Song-ICLR-2024,Song-ICML-2023,Luo-arXiv-2023,salimans-arXiv-2022,Yin-NeurIPS-2024,Yin-CVPR-2024,Sauer-ECCV-2024}. In such methods, a pre-trained student model is usually trained to predict the output of the teacher model over much larger denoising intervals. The resulting student is able to bypass a large portion, or even the entirety, of the intermediate denoising steps, enabling few-step or single-step image generation, while almost preserving the visual fidelity of the original model. Despite these optimizations achieved during inference, there are only a few works~\citep{dai-arXiv-2023,li-ICASSP-2025,li-BMVC-2025} that tackle training efficiency. However, these methods either require a manual selection~\citep{dai-arXiv-2023} of the most representative samples, or are tested only on labeled datasets because the selection process is individually performed for each class~\citep{li-ICASSP-2025}. This class-specific selection limits their applicability in real-world setups, like text-to-image generation. In contrast, we propose a fully-automated selection method and we show that it performs well on text-to-image synthesis.

In this work, we focus on reducing the computational overhead of fine-tuning diffusion models by selecting a diverse subset of exemplar images from the training data, as illustrated in Figure \ref{fig:placeholder123}. We take inspiration from the seminal works of \citep{Sirovich-JOSA-1987} and \citep{Pentland-CVPR-1994}, who apply Principal Component Analysis (PCA) to extract eigenfaces, i.e.\ a basis set of images that captures the greatest variance within a dataset of faces. However, the method is not robust to high data variations, e.g.\ natural images, as it is sensitive to affine transformations. To mitigate this problem, we propose to apply PCA in the CLIP latent space, to obtain a set of eigenvectors. We then use cosine similarity to retrieve the images that best align with the eigenvectors, ensuring that our subset spans the broad variations of the original data. However, selecting only the highest-aligning images fails to capture the internal diversity along each principal component. Therefore, we complement these initial exemplars using a stratified sampling strategy, selecting additional images that represent varying degrees of alignment to each component. Both dataset distillation steps are carried out with a focus on training budget, aiming to reduce the overall training time.

We conduct experiments on four datasets (Pick-a-Pic \citep{Kirstain-NeurIPS-2023}, HPDv2 \citep{Wu-arXiv-2023}, Rapidata \citep{christodoulou-arxiv-2024}, ROCO Radiology \citep{Pelka-IICASLABDELS-2018}), using two different diffusion models (Stable Diffusion 1.5~\citep{rombach-CVPR-2022} and PixArt-$\alpha$~\citep{Chen-ICLR-2024}). We consider two training setups, preference optimization and domain adaptation. Specifically, we verify the compatibility of our selection process with Direct Preference Optimization (DPO) \citep{Rafailov-NeurIPS-2023}, as well as supervised fine-tuning (SFT). Our results show that our dataset distillation method based on eigenvectors produces similar performance levels, while reducing training time by several folds (between $5.86\times$ and $13.95\times$).

In summary, our contribution is twofold:
\begin{itemize}
    \item We introduce EigDiff, a dataset distillation method that significantly reduces the computational overhead of fine-tuning diffusion model by selecting representative samples from the training set based on eigenvectors.
    \item We verify the effectiveness and efficiency of EigDiff on different datasets, diffusion models, and fine-tuning strategies.
\end{itemize}

\vspace{-0.2cm}
\section{\uppercase{Related Work}}
\label{sec:related_work}
\vspace{-0.2cm}

\noindent
\textbf{Diffusion models.} Although diffusion models have been introduced a while ago in~\citep{sohl-icml-2015}, it is only through the more recent denoising probabilistic diffusion models \citep{ho-NeurIPS-2020} and their score-based counterparts \citep{song-NeurIPS-2019,song-ICLR-2021} that they have established themselves as the new paradigm for generative image modeling \citep{Croitoru-TPAMI-2023}. Their recognition stemmed from greatly surpassing generative adversarial networks \citep{goodfellow-ACM-2020}, the previous leading paradigm, on both image fidelity and diversity \citep{dhariwal-NeurIPS-2021}. The scalability of diffusion models has been enabled through subsequent works \citep{bao-CVPR-2023,ho-NeurIPS-2021,Peebles-ICCV-2023,rombach-CVPR-2022}. Latent diffusion models \citep{rombach-CVPR-2022} reduced computational cost by operating in the compressed latent space of a pretrained autoencoder. \citep{nichol-ICML-2021} refined the noise schedule, while~\citep{Peebles-ICCV-2023} and~\citep{bao-CVPR-2023} leveraged more efficient architectures. 
Later, \citep{Song-ICML-2023} proposed a new type of diffusion models (called consistency models) that require only a few iterations. Nevertheless, all of these methods implicitly assume the availability of very large training corpora, a constraint we aim to mitigate in our work.

\noindent
\textbf{Time step distillation.} One of the main drawbacks of diffusion models is the large number of required sequential denoising steps, thus motivating a rich line of work on timestep distillation \citep{salimans-arXiv-2022,Yin-CVPR-2024,Yin-NeurIPS-2024,berthelot-arXiv-2023,gu-ICML-2023,Song-ICLR-2024,Song-ICML-2023,Luo-arXiv-2023,Frans-ICLR-2025,Sauer-ECCV-2024,meng-CVPR-2023,chadebec-AAAI-2025}. One of the earlier pieces of work, namely Progressive Distillation \citep{salimans-arXiv-2022}, aims to train a student model to match the output of two teacher steps in a single step, therefore halving the required number of steps. The most notable work in this direction is represented by Consistency Models \citep{Song-ICML-2023,Luo-arXiv-2023}. These aim few-step generation by enforcing self-consistency along the entire probability flow ODE trajectory, i.e.\ directly learning the mapping between noisy and clean samples across multiple timestamps. This enables single-step generation, while multi-step refinement remains a valid option for improved quality. While time step distillation can significantly reduce inference time, the full dataset and the full model are still required during training, thereby limiting model fine-tuning with typical academic resources.

\noindent
\textbf{Architecture distillation.} Another adopted method for reducing the computational complexity and the total parameter count of diffusion models, while preserving the generation quality, is based on architectural distillation \citep{kim-ECCV-2024,zhou-ICML-2024,moon-ICML-2024}, i.e.~compressing the architecture. Modern diffusion models rely on heavily over-parameterized U-Nets with attention mechanisms or transformer-based backbones. As a result, on the one hand, several works have investigated transferring knowledge from large teacher networks to lighter student architectures through feature matching or score alignment \citep{kim-ECCV-2024,zhou-ICML-2024,xiang-CVPR-2025,lee-NeurIPS-2024}, and, on the other hand, the models are pruned \citep{fang-NeurIPS-2023,zhang-arXiv-2024,Yu-TMM-2025,gupta-arXiv-2024,yuan-NeurIPS-2024}. \citep{kim-ECCV-2024} proposed to reduce the blocks from the U-Net by employing three objectives: estimate the added noise during the forward pass (the usual objective of diffusion models), estimate the predicted noise of the teacher model, and match the feature maps between the teacher and student models. They managed to reduce the model size up to $51\%$, while improving the latency by $43\%$. \citep{lee-NeurIPS-2024} scaled this approach to SDXL. \citep{fang-NeurIPS-2023} presented a distillation method that offers a trade-off between efficiency and quality: sub-structures from the network are removed based on some importance scores computed using the gradient. Subsequent works focus on eliminating the redundancy of different parts from a diffusion model: layer-level distillation \citep{zhang-arXiv-2024,gupta-arXiv-2024}, attention reduction \citep{yuan-NeurIPS-2024}, and even combining distillation and pruning \citep{zhang-arXiv-2024,Yu-TMM-2025}. While effective, architecture distillation requires inference through the teacher model, which can represent a significant overhead during training. Our contribution is focused on dataset distillation, an orthogonal direction focused on reducing training time.

\noindent
\textbf{Dataset distillation.} Deep learning datasets often contain substantial redundancy, in which the contribution of many samples bring limited information gains. Several studies \citep{cazenavette-CVPR-2022,Hondru-AIR-2025,yu-TPAMI-2024,wang-arXiv-2018,zhao-ICLR-2021} have shown that large neural networks can maintain strong performance even when trained on carefully selected subsets of the original data, although these have mainly been explored almost exclusively in the supervised classification regime. \citep{dai-arXiv-2023} analyzed the impact of carefully selecting the training data for a text-to-image diffusion model and conclude that fine-tuning on a very small yet highly visually appealing set of images can substantially enhance generation quality. However, unlike our approach, after an initial automatic filtering (offensive content removal, aesthetic and CLIP-score filtering, OCR-based text filtering, image size and aspect ratio constraints, balancing categories such as portraits, landscapes, animals, food images, and finally some proprietary quality checks), they employed a manual labeling process in order to select candidate images. More recently, \citep{li-ICASSP-2025} proposed a scheme for pruning the training data. Embeddings of the data samples are initially obtained using the latent space of a pre-trained image classifier, and then, the most informative examples are selected according to a Gaussian-based model scoring criterion. After this, class weights are dynamically computed through distributionally robust optimization, using a pre-trained reference diffusion model. In contrast to existing methods, that either employ manual labeling \citep{dai-arXiv-2023} or require a computationally intensive selection process involving deep neural models \citep{li-ICASSP-2025}, we propose a lightweight dataset distillation procedure that selects samples based on their similarity to a set of eigenvectors computed in a CLIP latent space. 

\vspace{-0.2cm}
\section{\uppercase{Method}}
\label{sec:method}
\vspace{-0.2cm}

\noindent
\textbf{Motivation.} EigDiff aims to identify a compressed training set of representative images that preserve (i) the dominant characteristics and (ii) the diversity of the entire dataset, while reducing data redundancy by as much as possible. To achieve these goals, we first extract EVA-CLIP~\citep{sun2023evaclipimprovedtrainingtechniques} embeddings from the images and apply Principal Component Analysis (PCA) to compute eigenvectors, which serve as a basis for reconstructing the embedding space. This approach is inspired by eigenfaces \citep{Pentland-CVPR-1994,Sirovich-JOSA-1987}. However, because we operate on more complex visual domains, we cannot use raw eigenimages directly as training data for generative models. The eigenimages are highly sensitive to lighting variations and affine transformations, and they lack the high-frequency details specific to natural images, which are required for training high-quality generative models. To achieve (i), we select exemplars that encapsulate the dominant characteristics. We accomplish this by locating the latent vectors closest to each eigenvector within the embedding space and retaining its corresponding natural image. To achieve the second goal (ii), we propose to select additional samples using a stratified sampling approach, where the stratification is applied via similarity bins computed w.r.t.~the eigenvectors. Selecting images from different similarity bins ensures a broad diversity among the select samples. 

\noindent
\textbf{Data preprocessing.}
We evaluate our method under two training paradigms: supervised fine-tuning (SFT) and direct preference optimization (DPO). For SFT setup, the dataset comprises $N$ image-text pairs, $D_{\text{raw}} = \{(I_i, c_i)\}_{i=1}^N$, where $I_i \in \mathbb{R}^{H \times W \times C}$ represents an image of resolution $H \times W$ with $C$ channels, and $c_i$ is a corresponding text prompt. For the DPO setup, the dataset is a collection of $N$ triplets, $D_{\text{raw}} = \{(I_i^w, I_i^l, c_i)\}_{i=1}^N$, where $I_i^w, I_i^l \in \mathbb{R}^{H \times W \times C}$ denote the chosen (winning) and rejected (losing) images, respectively, both generated from the same prompt $c_i$.

Let $g : \mathbb{R}^{H \times W \times C} \to \mathbb{R}^d$ be a frozen image encoder that projects the images into a $d$-dimensional space. In our case, the encoder is EVA-CLIP~\citep{sun2023evaclipimprovedtrainingtechniques}. In the preference optimization setup, we obtain $L_2$-normalized embeddings for the chosen and rejected images as $x_i^w = \frac{g(I_i^w)}{\lVert g(I_i^w)\rVert}$ and $x_i^l = \frac{g(I_i^l)}{\lVert g(I_i^l)\rVert}$. Stacking these representations, we obtain the chosen and rejected embedding matrices, $X^w = [x_1^w, \dots, x_N^w]^\top \in \mathbb{R}^{N \times d}$ and $X^l = [x_1^l, \dots, x_N^l]^\top \in \mathbb{R}^{N \times d}$, respectively. Following an identical approach for SFT, we compute normalized embeddings for the individual images and we obtain a similar embedding matrix, $X = [x_1, \dots, x_N]^\top \in \mathbb{R}^{N \times d}$.

\noindent
\textbf{Principal components selection.} We fit PCA on $X^w$ for DPO, and on $X$ for SFT. We use only $X^w$ for DPO because we want the recovered principal components to describe the dominant modes of variation of the preferred images. After eigen decomposition of the sample covariance, we obtain $m$ eigenvectors $W = \left[w_1\ \dots\ w_m\right]^\top \in \mathbb{R}^{m \times d}$ sorted by their corresponding eigenvalues $(\lambda_i)_{i=1}^m$, which express the amount of data variance explained by the direction of each eigenvector. We limit the number of principal components used for sample selection by selecting the top principal components based on the explained variance. To achieve this, we compute the ratios of explained variance and the cumulative-variance, as follows:
\begin{equation}
    \hat{
    \lambda}_i = \frac{\lambda_i}{\sum_{j=1}^{m}\lambda_j},
    \qquad
    \bar{\lambda}_i = \sum_{j=1}^{i}\hat{\lambda}_j, i \in \{1, \dots, m\},
\end{equation}
where $\hat{\lambda}_i$ represents the ratio of explained variance of the $i$-th principal component, and $ \bar{\lambda}_i$ represents the total variance explained by the first $i$ principal components.



We retain components up to two cumulative-variance thresholds $\tau_1 <\tau_2$. Based on these thresholds, we partition the principal components into two disjoint subsets. We use the first set to select exemplars and the second to enhance sample diversity. These subsets are determined as follows:
\begin{align}
    \mathcal{A}&=\{i\mid\bar{\lambda}_i\leq\tau_1,\ i=\overline{1,m}\},\nonumber\\
    \mathcal{B}&=\{i\mid\tau_1<\bar{\lambda}_i\leq\tau_2,\ i=\overline{1,m}\}.\label{eq:pc_quantiles}
\end{align}
 
\noindent
Essentially, the first set $\mathcal{A}$ carries the strongest variance directions, yielding the non-repeating exemplars, whilst the diversity set $\mathcal{B}$ carries the secondary structure which is, by design, disjoint from the other. 

\noindent
\textbf{Stage 1.}
\label{sec:stage1}
In the first stage, we leverage the principal components from $\mathcal{A}$ to select the exemplars. For this, we fix a budget of $n_1$ samples that we consume based on the explained variance of each principal component. Specifically, for the $i$-th principal component from $\mathcal{A}$, we select as exemplars the most similar $k_i$ samples from the dataset. We establish $k_i$ as follows:
\begin{equation}
    k_i \;=\; \left\lfloor \frac{\hat{\lambda}_i}{\sum_{j\in\mathcal{A}}\hat{\lambda}_j}\, n_1\right\rfloor,
    \forall i\in\mathcal{A}.
    \label{eq:variance}
\end{equation}

We decide which samples are the most similar with a given principal component via cosine similarity, as follows:
\begin{equation}
\label{eq:scores}
    \!\!\!\!S_{i,j}\!=\!
    \begin{cases}
       w_i(x_j^w - x_j^l)^{\top}\!\!\!\!\! & \text{(DPO)}\\
        w_ix_j^{\top} & \text{(SFT)}
    \end{cases}, \forall i\in \mathcal{A}, j=\overline{1, N}.
\end{equation}
In the DPO case, $S_{i, j}$ is a signed preference contrast: it is large and
positive when the chosen image embedding aligns with $w_i$ more strongly than the
rejected one, near zero when the pair is undecided along $w_i$, and negative
when the contrast is reversed. The sign is retained throughout and encodes the
preference direction which the downstream objective consumes. In the SFT case, the score
is the raw alignment of the image embedding with $w_i$.

We define $\mathcal{S}_i^1$ as the set containing the $k_i$ samples most similar to the $i$-th principal component, based on the similarity scores computed in $S_i$. These samples consist of triplets $(I^w, I^l, c)$ for DPO, and pairs $(I, c)$ for SFT. Consequently, the final dataset utilized during Stage 1 training is given by:
\begin{equation}
    \mathcal{S}^1 = \bigcup_{i\in\mathcal{A}}\mathcal{S}_i^1.
\end{equation}

\noindent\textbf{Stage 2.} In the second stage, we leverage the principal components in $\mathcal{B}$ to construct a diverse sample set. Following the approach from Stage 1, we define a total sample budget $n_2 = \delta \cdot n_1$, allocating $k_i$ samples to each principal component as in Eq.~\eqref{eq:variance}. For every diversity component index $i \in \mathcal{B}$, we compute the similarity scores $S_{i,j}$ (where $j \in \{1, \dots, N\}$), mirroring the procedure in Eq.~\eqref{eq:scores}. Next, we partition the score distribution $S_i$ of each component into $B$ evenly-spaced empirical quantile intervals, where $B \in \mathbb{N}$ and $B \ge 1$. The limits of these intervals are defined as:
\begin{gather}
\!\!\!l_b\!=\!\mathrm{quantile}\!\left(\{S_{i,j}\}_{j=1}^N, \tfrac{b}{B}\right), \forall b \in \{0,\dots,B\!-\!1\}.
    \label{eq:nobins}
\end{gather}
For each principal component $i \in \mathcal{B}$ and interval $b$, we uniformly draw $\lfloor k_i / B \rfloor$ samples:
\begin{multline}\label{eq:draw}
\mathcal{S}^2_{i,b}\sim\mathcal{U}\Big(\{(I^w_j,I^l_j,c_j)\mid l_b\leq S_{i,j}<l_{b+1},\\
j=1,\dots,N\},\left\lfloor\frac{k_i}{B}\right\rfloor\Big),
\end{multline}
where $\mathcal{U}(A, z)$ denotes uniformly sampling $z$ elements from set $A$. The aggregate sample set generated in this stage is given by:
\begin{equation}
    \mathcal{S}^2 = \bigcup_{i\in\mathcal{B}}\bigcup_{b=0}^{B-1}\mathcal{S}_{i,b}^2.
\end{equation}
Ultimately, the complete training set is formed by the union of the subsets from both stages:
\begin{equation}
    \mathcal{S} = \mathcal{S}^1 \cup \mathcal{S}^2.
    \vspace{-0.3cm}
\end{equation}

\begin{table*}[!t]
\caption{Per-dataset training steps for the full-scale baseline and our
eigen-based selection method. Additionally, we include the hyperparameters of our method.}
\vspace{-0.2cm}
\begin{center}
\setlength{\tabcolsep}{4pt}
  \fontsize{8}{9}\selectfont{
\begin{tabular}{| l | l | c | c | c | c | c | c |}
\hline
\multirow{2}{*}{\textbf{Dataset}} & \multirow{2}{*}{\textbf{Backbone}}
 & \textbf{Full-scale} & \textbf{Eigen (ours)}
 & \multirow{2}{*}{$n_1$} & \multirow{2}{*}{$\delta$} & \multirow{2}{*}{$B$} & \multirow{2}{*}{$(\tau_1,\tau_2)$} \\
\cline{3-4}
 & & Steps & Steps & & & & \\
\hline
\multirow{2}{*}{Pick-a-Pic}
 & SD1.5  & $50{,}000$ & $3{,}584$ & \multirow{2}{*}{$1{,}024$} &  \multirow{2}{*}{$6$} &  \multirow{2}{*}{$4$} & \multirow{2}{*}{$(0.6,0.8)$} \\
 & PixArt-$\alpha$ & $25{,}000$ & $1{,}792$ & & & &\\
\hline
\multirow{2}{*}{HPDv2}
 & SD1.5  & $50{,}000$ & $7{,}168$ & \multirow{2}{*}{$2{,}048$} & \multirow{2}{*}{$6$}  &  \multirow{2}{*}{$4$}  & \multirow{2}{*}{$(0.6,0.8)$} \\
 & PixArt-$\alpha$ & $25{,}000$ & $3{,}584$ & & & & \\
\hline
\multirow{2}{*}{Rapidata}
 & SD1.5  & $22{,}515$ & $3{,}584$ & \multirow{2}{*}{$1{,}024$} &  \multirow{2}{*}{$6$}  &  \multirow{2}{*}{$4$}  &\multirow{2}{*}{$(0.4,0.6)$} \\
 & PixArt-$\alpha$ & $11{,}257$ & $1{,}792$ & & & & \\
\hline
\multirow{2}{*}{ROCO}
 & SD1.5  & $56{,}066$ & $9{,}557$ & \multirow{2}{*}{$1{,}024$} &  \multirow{2}{*}{$6$} &  \multirow{2}{*}{$4$} & \multirow{2}{*}{$(0.4,0.6)$} \\
 & PixArt-$\alpha$ & $84{,}100$ & $14{,}336$ & & & &\\
\hline
\end{tabular}
}
\end{center}
\label{tab:dataset_schedules}
\vspace{-0.2cm}
\end{table*}

\vspace{-0.2cm}
\section{\uppercase{Experiments}}
\label{sec:experiments}
\vspace{-0.2cm}
\subsection{Datasets}
\vspace{-0.2cm}

We evaluate EigDiff on one medical text-to-image dataset, ROCO Radiology~\citep{Pelka-IICASLABDELS-2018}, alongside three text-to-image preference datasets: Pick-a-Pic v1~\citep{Kirstain-NeurIPS-2023}, HPDv2~\citep{Wu-arXiv-2023}, and Rapidata~\citep{christodoulou-arxiv-2024}. For all preference datasets, we first filter out the non-decision samples. To manage training size, we randomly subsample Pick-a-Pic v1 and HPDv2 to $100,000$ training examples each, while retaining their original test sets of $500$ and $400$ prompts, respectively. Finally, from Rapidata, we utilize the human style preference split, allocating $45,030$ pairs for training and $75$ unique prompts for testing.

ROCO Radiology \citep{Pelka-IICASLABDELS-2018} consists of pairs of medical images and captions, and it serves as an out-of-domain dataset for our selected generators. We apply two data filtering operations: retaining only the training samples with an aspect ratio between $0.7$ and $1.35$ and discarding samples with captions shorter than 6 words. Consequently, we remain with $42,050$ preference pairs for training, and we utilize $500$ prompts for testing.


\vspace{-0.2cm}
\subsection{Training Setup}
\vspace{-0.2cm}
\noindent
\textbf{Baselines.} We compare our approach against seven baselines. The first baseline consists of the pre-trained models, Stable Diffusion 1.5~\citep{rombach-CVPR-2022} (SD1.5) and PixArt-$\alpha$-XL-2~\citep{Chen-ICLR-2024} (PixArt-$\alpha$). The second is a full-scale fine-tuning on the entire datasets, representing a performance target that our method aims to match. Finally, we compare against a random selection variant, which keeps the exact same number of training samples as our method, but selects them randomly. In addition, we compare against four data selection methods: Greedy K-Center~\citep{Gonzalez-TCS-1985}, Stochastic-Greedy Facility Location~\citep{Mirzasoleiman-AAAI-2015}, K-Means++~\citep{Arthur-SODA-2007}, and SemDeDup~\citep{Abbas-arXiv-2023}. They follow the same configuration and budget ($n_1+n_2$) as our approach. 
In the case of RapiData, methods such as K-Center, K-Means++ and SemDeDup retain all 4,888 unique samples, thus we train them for 2,444 steps with SD1.5, and 1,222 steps with PixArt-$\alpha$.


\noindent
\textbf{Hyperparameters.} For both backbones, we employ AdamW with $\beta_1=0.9$, $\beta_2=0.999$, weight decay equal to $10^{-2}$, $\epsilon=10^{-8}$, with maximum gradient clipping at norm $1.0$. We use a training batch size of $6$ for SD1.5 and a batch size of $4$ for PixArt-$\alpha$. We set the learning rate to $10^{-5}$. The learning rate schedule is constant and it has a warm-up phase.  We use the classifier-free guidance with $20\%$ empty prompts. In order to reduce the computational complexity, we train the models using LoRA \citep{Hu-ICLR-2022}. For our LoRA configuration, we use a rank of $16$, $\alpha = 16$, and dropout $0.01$. The LoRA layers are applied on the query, key, value and output projections layers, 
resulting in PixArt-$\alpha$ having $1.33\%$ trainable parameters and SD1.5 $0.37\%$. On Pick-a-Pic, HPDv2, and Rapidata, we train SD1.5 for $3$ epochs, and PixArt-$\alpha$ for $1$ epoch. On ROCO Radiology, we fine-tune both models for $8$ epochs. In the DPO setup, we set  $\beta=2000$ for SD1.5 and  $\beta=5000$ for PixArt-$\alpha$.

At inference time, we use $50$ denoising steps with classifier-free guidance of scale $7.5$ for SD1.5, and $20$ denoising steps with $4.5$ scale for PixArt-$\alpha$.

In Table~\ref{tab:dataset_schedules}, we detail the hyperparameters specific to EigDiff. Specifically, for each dataset, we present the budget of samples for the first stage $n_1$, the ratio of samples for the second stage $\delta=\frac{n_2}{n_1}$, the number of intervals $B$ employed in second stage, and the variance thresholds $(\tau_1, \tau_2)$. Additionally, we showcase the difference in total training steps between our approach and the full-scale fine-tuning baseline. Note that these step counts reflect the total number of iterations required to reach the previously specified number of epochs.

\begin{table*}[!t]
\caption{Total fine-tuning time (in minutes) for full-scale vs.~distilled datasets. Experiments are performed with an equivalent number of epochs, on one Intel i9-14900F CPU with 64GB RAM and an NVIDIA RTX 5090 GPU. Less time is better.}
\vspace{-0.2cm}
\begin{center}
  \fontsize{8}{9}\selectfont{
\begin{tabular}{|l|l|c|c|c|c|c|c|}
\hline
\multirow{2}{*}{\textbf{Objective}} &
\multirow{2}{*}{\textbf{Dataset}} &
\multicolumn{3}{c|}{\textbf{SD1.5}} &
\multicolumn{3}{c|}{\textbf{PixArt-$\alpha$}} \\
\cline{3-8}
&
& \textbf{Full-scale} & \textbf{EigDiff (ours)} & \textbf{Speedup}
& \textbf{Full-scale} & \textbf{EigDiff (ours)} & \textbf{Speedup} \\
\hline

\multirow{3}{*}{DPO}
& Pick-a-Pic & 407 & 29.175 & 13.95$\times$ & 184.26  & 13.208 & 13.95$\times$ \\
& HPDv2      & 407 &  58.35
 & 6.98$\times$& 184.26 & 26.416 & 6.98$\times$ \\
& Rapidata   & 183.27 & 29.175 & 6.28$\times$ & 82.97 & 13.207 & 6.28$\times$ \\
\hline

\multirow{4}{*}{SFT}
& Pick-a-Pic &  183.13 & 13.122  & 13.95$\times$  & 97.20 & 6.97 &13.95$\times$ \\
& HPDv2      & 183.13 & 26.244 & 6.98$\times$  & 97.20 & 13.94 & 6.98$\times$ \\
& Rapidata   & 82.47 & 13.127 &6.28$\times$  & 43.77 & 6.97 &6.28$\times$  \\
& ROCO       & 205.35 & 34.947 &5.87$\times$  & 326.98 &  55.76 &5.87$\times$  \\
\hline

\end{tabular}
}
\end{center}
\label{tab:compute}
\end{table*}

\begin{table*}[!t]
\caption{Average selection time comparison (in seconds) over three seeds with standard deviation. Selections run on their respective pre-computed features, thus the timing is identical regardless of the model. We showcase the speedup interval between EigDiff, and the slowest/fastest method (SemDeDup to Facility Locator). We used an Intel i9-14900F CPU with 64GB of RAM for all selections, each approach being measured with 4 threads. Less time is better.}
\vspace{-0.2cm}
\begin{center}
  \fontsize{8}{9}\selectfont{
\begin{tabular}{|l|l|c|c|c|c|c|c|}
\hline
\textbf{Objective} & \textbf{Dataset} & \textbf{K-Center} & \textbf{Facility Loc.} & \textbf{K-Means$++$}
& \textbf{SemDeDup} & \textbf{EigDiff (ours)} & \textbf{Speedup} \\
\hline

\multirow{3}{*}{DPO}
& Pick-a-Pic & $358.1_{\pm0.3}$ & $1525.5_{\pm18.0}$ & $360.6_{\pm0.7}$ & $5.27_{\pm0.24}$ & $1.36_{\pm0.02}$ &
3.9--1123$\times$ \\
& HPDv2      & $708.2_{\pm6.0}$ & $2079.8_{\pm43.0}$ & $715.9_{\pm5.7}$ & $4.48_{\pm0.21}$ & $1.46_{\pm0.03}$ &
3.1--1424$\times$ \\
& Rapidata   & $106.6_{\pm1.6}$ & $468.5_{\pm0.1}$ & $110.8_{\pm0.8}$ & $2.06_{\pm0.29}$ & $0.46_{\pm0.01}$ &
4.4--1011$\times$ \\
\hline

\multirow{4}{*}{SFT}
& Pick-a-Pic & $358.1_{\pm0.3}$ & $1525.5_{\pm18.0}$ & $360.6_{\pm0.7}$ & $5.27_{\pm0.24}$ & $1.23_{\pm0.04}$ &
4.3--1244$\times$ \\
& HPDv2      & $708.2_{\pm6.0}$ & $2079.8_{\pm43.0}$ & $715.9_{\pm5.7}$ & $4.48_{\pm0.21}$ & $1.30_{\pm0.02}$ &
3.4--1598$\times$ \\
& Rapidata   & $106.6_{\pm1.6}$ & $468.5_{\pm0.1}$ & $110.8_{\pm0.8}$ & $2.06_{\pm0.29}$ & $0.38_{\pm0.02}$ &
5.4--1230$\times$ \\
& ROCO       & $150.8_{\pm1.6}$ & $447.2_{\pm4.0}$ & $152.4_{\pm1.2}$ & $2.05_{\pm0.19}$ & $0.28_{\pm0.01}$ &
7.4--1607$\times$ \\
\hline

\end{tabular}
}
\end{center}
\label{tab:selection_time}
\vspace{-0.2cm}
\end{table*}
\noindent
\textbf{Evaluation metrics.} For the in-domain sets (Pick-a-Pic, HPDv2, Rapidata), we report PickScore \citep{Kirstain-NeurIPS-2023}, a CLIP-based scoring function that indicates the human-preference alignment between generated images and their prompts. Since PickScore is a general domain predictor, it is not calibrated for medical images. As a result, since ROCO Radiology is a medical corpus, we report the BiomedCLIP score \citep{zhang2024biomedclip}, which measures the alignment in the clinical domain.



\vspace{-0.2cm}
\subsection{Quantitative Results}
\vspace{-0.2cm}

\noindent
\textbf{Computational gains.}
We showcase the time improvement of our approach compared with the full-scale regime in Table \ref{tab:compute}. We fine-tune all models for the same number of epochs, resulting in a $5.87-13.95\times$ fine-tuning speedup with our dataset distillation approach compared with the full-scale runs, based on the size of the selected coresets. In Table \ref{tab:selection_time}, we compare the speeds of various coreset selection approaches. Among all methods, our eigen-based approach yields significantly faster runtime, being by $3.1$-$1607\times$ faster than other methods. All baselines operate only on the features of preferred images, motivating why timings are identical between DPO and SFT. Conversely, our approach leverages both preferred and non-preferred images, in the case of DPO (Eq.~\eqref{eq:scores}). We run our experiments on an Nvidia RTX 5090 with 32GB of VRAM, $16\times4$ GB of RAM, and an Intel Core I9-14900F CPU with 24 cores. 

\begin{figure}[!t]
    \centering
    \includegraphics[width=1.0\linewidth]{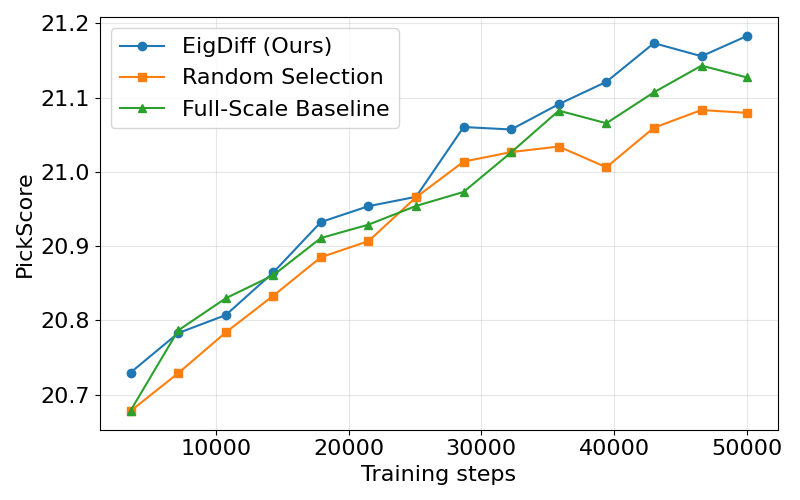}
    \caption{Comparing the experiments with Stable Diffusion 1.5 on Pick-a-Pic, fine-tuned on 50000 steps, on the best performing full-scale seed.}
    \label{fig:placeholder}
\end{figure}

\noindent
\textbf{DPO fine-tuning.} 
We showcase the performance of EigDiff vs.~baselines in the DPO regime in Table \ref{tab:results_DPO}. Both reference points (pre-trained and full-scale) are strong here. The full-scale runs are fine-tuned using the same number of epochs as the selection methods, thus taking $6.3$-$14\times$ more training steps (see Table \ref{tab:dataset_schedules}). The pre-trained schedule already scores high on all sets, as their samples are similar to those from the training corpora of SD1.5 and PixArt-$\alpha$, therefore leaving little room for improvement. Even so, EigDiff beats random selection in all cases. Our approach is best in four cases, and second-best for PixArt-$\alpha$ on Pick-a-Pic and HPDv2. The EigDiff with SD1.5 leads on all data sets, with the largest gain over random being on Pick-a-Pic ($+0.036$). Here as well, full-scale is furthest ahead, though Figure \ref{fig:placeholder} indicates that giving EigDiff the same 50{,}000 steps as full-scale, it overtakes full-scale in later steps. With PixArt-$\alpha$, our approach comes close to Facility Locator by 0.003 on Pick-a-Pic, and SemDeDup by 0.029 on HPDv2. EigDiff exceeds the full-scale run in three settings: on RapiData by 0.024 with SD1.5, by 0.007 with PixArt-$\alpha$, and on HPDv2 with PixArt-$\alpha$ by 0.034. In this case, every fine-tuned method stays below the pre-trained strategy. On Pick-a-Pic with PixArt-$\alpha$, EigDiff is within 0.001 of the full-scale run. On DPO, the selection step of eigen is also $3.1$-$1424\times$ faster than the other selection strategies (see Table \ref{tab:selection_time}). 

\begin{table*}[!t]
\caption{PickScore results (higher is better) using SD1.5 and PixArt-$\alpha$ under the DPO regime, on the Pick-a-Pic, HPDv2, and Rapidata datasets, comparing our approach with the baselines. We report the average PickScore over three runs. For model and dataset, we highlight the best and second-best scores among coreset selection / dataset distillation strategies in bold and underline, respectively. The pre-trained and full-scale experiments are reference points, thus not ranked.}
\vspace{-0.2cm}
\begin{center}
\setlength{\tabcolsep}{5pt}
\fontsize{8}{9}\selectfont{
\begin{tabular}{|l|l|c|c|c|}
\hline
\textbf{Model} &
\textbf{Strategy} &
\textbf{Pick-a-Pic} &
\textbf{HPDv2} &
\textbf{Rapidata} \\
\hline

\multirow{9}{*}{SD1.5}
& Pre-trained
& $20.536$
& $20.779$
& $20.373$ \\

\cline{2-5}

& K-Means$++$~\citep{Arthur-SODA-2007}
& $20.691_{\pm0.004}$
& $\underline{20.989}_{\pm0.014}$
& $20.390_{\pm0.013}$ \\

& K-Center~\citep{Gonzalez-TCS-1985}
& $20.645_{\pm0.008}$
& $20.571_{\pm0.030}$
& $20.394_{\pm0.009}$ \\

& Facility Location~\citep{Mirzasoleiman-AAAI-2015}
& $\underline{20.702}_{\pm0.010}$
& $20.983_{\pm0.065}$
& $20.414_{\pm0.028}$ \\

& SemDeDup~\citep{Abbas-arXiv-2023}
& $20.614_{\pm0.013}$
& $20.898_{\pm0.029}$
& $20.398_{\pm0.005}$ \\

& Random selection
& $20.669_{\pm0.013}$
& $20.977_{\pm0.016}$
& $20.432_{\pm0.022}$ \\

& EigDiff (no EVA-CLIP)
& $20.689_{ \pm0.005}$
& $20.986_{ \pm0.031}$
& $\underline{20.447}_{ \pm0.020}$ \\

& EigDiff (ours)
& $\textbf{20.705}_{\pm0.023}$
& $\textbf{20.994}_{\pm0.021}$
& $\textbf{20.459}_{\pm0.013}$ \\

\cline{2-5}

& Full-scale dataset
& $21.107_{\pm0.018}$
& $21.019_{\pm0.021}$
& $20.435_{\pm0.013}$ \\

\hline

\multirow{9}{*}{PixArt-$\alpha$}
& Pre-trained
& $21.979$
& $22.517$
& $21.553$ \\

\cline{2-5}

& K-Means$++$~\citep{Arthur-SODA-2007}
& $21.984_{\pm0.004}$
& $22.456_{\pm0.021}$
& $21.549_{\pm0.009}$ \\

& K-Center~\citep{Gonzalez-TCS-1985}
& $21.982_{\pm0.002}$ 
& $22.395_{\pm0.022}$
& $\underline{21.550}_{\pm0.010}$ \\

& Facility Location~\citep{Mirzasoleiman-AAAI-2015}
& $\textbf{21.992}_{\pm0.005}$
& $22.415_{\pm0.012}$ 
& $21.548_{\pm0.016}$ \\

& SemDeDup~\citep{Abbas-arXiv-2023}
& $21.982_{\pm0.004}$ 
& $\textbf{22.492}_{\pm0.026}$
& $21.548_{\pm0.012}$ \\

& Random selection
& $21.983_{\pm0.006}$
& $22.442_{\pm0.028}$
& $21.545_{\pm0.011}$ \\

& EigDiff (no EVA-CLIP)
& $21.988_{ \pm0.002}$
& $22.449_{ \pm0.022}$
& $21.535_{ \pm0.005}$ \\

& EigDiff (ours)
& $\underline{21.989}_{\pm0.003}$
& $\underline{22.463}_{\pm0.021}$
& $\textbf{21.561}_{\pm0.005}$ \\

\cline{2-5}
& Full-scale dataset
& $21.990_{\pm0.006}$
& $22.429_{\pm0.012}$
& $21.554_{\pm0.042}$ \\

\hline
\end{tabular}
}
\end{center}
\label{tab:results_DPO}
\vspace{-0.1cm}
\end{table*}

\begin{table*}[!t]
\caption{PickScore/BiomedCLIP results (higher is better) using SD1.5 and PixArt-$\alpha$ under the SFT regime, on the Pick-a-Pic, HPDv2, Rapidata and ROCO datasets, comparing our approach with the baselines. We report the average PickScore over three runs. For model and dataset, we highlight the best and second-best scores among coreset selection / dataset distillation strategies in bold and underline, respectively. The pre-trained and full-scale experiments are reference points, thus not ranked.}
\vspace{-0.2cm}
\begin{center}
\setlength{\tabcolsep}{4pt}
\fontsize{8}{9}\selectfont{
\begin{tabular}{|l|l|c|c|c|c|}
\hline
\textbf{Model} &
\textbf{Strategy} &
\textbf{Pick-a-Pic} &
\textbf{HPDv2} &
\textbf{Rapidata} &
\textbf{ROCO} \\
\hline

\multirow{4}{*}{SD1.5}
& Pre-trained
& $20.536$
& $20.779$
& $20.373$
& $15.4887$ \\

\cline{2-6}

& K-Means$++$~\citep{Arthur-SODA-2007}
& $20.808_{\pm0.005}$
& $20.777_{\pm0.009}$
& $20.561_{\pm0.013}$
& $21.031_{\pm0.124}$ \\

& K-Center~\citep{Gonzalez-TCS-1985}
& $20.778_{\pm0.026}$
& $20.721_{\pm0.019}$
& $20.529_{\pm0.006}$
& $21.098_{\pm0.225}$ \\

& Fac. Loc.~\citep{Mirzasoleiman-AAAI-2015}
& $\underline{20.820}_{\pm0.014}$
& $\underline{20.791}_{\pm0.023}$
& $\underline{20.579}_{\pm0.019}$
& $20.814_{\pm0.368}$ \\

& SemDeDup~\citep{Abbas-arXiv-2023}
& $20.718_{\pm0.005}$
& $20.786_{\pm0.020}$
& $20.521_{\pm0.042}$
& $20.375_{\pm0.061}$ \\

& Random selection
& $20.811_{\pm0.011}$
& $20.773_{\pm0.002}$
& $20.566_{ \pm0.016}$
& $\underline{21.184}_{\pm0.186}$ \\

& EigDiff (no EVA-CLIP)
& $20.808_{ \pm0.004}$
& $20.780_{ \pm0.020}$
& $20.575_{ \pm0.058}$
& $20.753_{\pm0.425}$ \\

& EigDiff (ours)
& $\textbf{20.825}_{ \pm0.014}$
& $\textbf{20.795}_{ \pm0.019}$
& $\textbf{20.582}_{ \pm0.040}$
& $\textbf{21.237}_{\pm0.293}$ \\
\cline{2-6}

& Full-scale dataset
& $20.994_{\pm0.006}$
& $20.781_{\pm0.004}$
& $20.848_{\pm0.018}$
& $24.526_{\pm0.365}$ \\

\hline

\multirow{9}{*}{PixArt-$\alpha$}
& Pre-trained
& $21.979$
& $22.517$
& $21.553$
& $15.452$ \\

\cline{2-6}

& K-Means$++$~\citep{Arthur-SODA-2007}
& $21.814_{\pm0.006}$ 
& $21.573_{\pm0.021}$
& $21.513_{\pm0.017}$
& $\underline{20.539}_{\pm0.126}$ \\

& K-Center~\citep{Gonzalez-TCS-1985}
& $21.827_{\pm0.007}$
& $21.551_{\pm0.039}$
& $21.509_{\pm0.008}$
& $20.076_{\pm0.204}$ \\

& Fac. Loc.~\citep{Mirzasoleiman-AAAI-2015}
& $\underline{21.830}_{\pm0.008}$
& $\textbf{21.675}_{\pm0.025}$
& $21.528_{\pm0.008}$
& $20.452_{\pm0.488}$ \\

& SemDeDup~\citep{Abbas-arXiv-2023}
& $\textbf{21.844}_{\pm0.006}$
& $\underline{21.662}_{\pm0.012}$
& $21.518_{\pm0.007}$
& $19.258_{\pm0.314}$ \\

& Random selection
& $ 21.814_{ \pm0.007}$
& $21.604_{ \pm0.038}$
& $\underline{21.531}_{ \pm0.010}$
& $20.491_{\pm0.321}$ \\

& EigDiff (no EVA-CLIP)
& $21.819_{ \pm0.006}$
& $21.607_{ \pm0.035}$
& $21.526_{ \pm0.008}$
& $20.295_{\pm0.328}$ \\

& EigDiff (ours)
& $21.821_{ \pm0.005}$
& $21.618_{ \pm0.037}$
& $\textbf{21.539}_{ \pm0.003}$
& $\textbf{20.916}_{\pm 0.329}$ \\

\cline{2-6}

& Full-scale dataset
& $	21.436_{ \pm0.026}$
& $	20.985_{ \pm0.042}$
& $	21.405_{ \pm0.021}$
& $24.272_{\pm 0.072}$ \\

\hline
\end{tabular}
}
\end{center}
\label{tab:results_SFT}
\vspace{-0.2cm}
\end{table*}


\noindent
\textbf{SFT fine-tuning.}
We present results with SFT in Table \ref{tab:results_SFT}. EigDiff again beats the random selection in all cases. With SD1.5, our approach is the best selection strategy on all four data sets and even exceeds the full-scale run on HPDv2. Full-scale is ahead in the other cases, having extra data. With PixArt-$\alpha$, no fine-tuned strategy reaches the pre-trained scores on Pick-a-Pic, HPDv2, and Rapidata, the in-domain data sets. In this case, the full-scale runs fall behind all selection approaches, which we attribute to catastrophic forgetting: with $6.3$-$14\times$ more steps, the model drifts furthest from its strong pre-trained weights, towards lower-quality fine-tuning data. With the PixArt-$\alpha$ backbone, EigDiff leads on Rapidata and ROCO, while K-Center, Facility Locator and SemDeDup score higher on Pick-a-Pic, with Facility Locator and SemDeDup leading on HPDv2. Since ROCO is out-of-domain, pre-trained models show low BiomedCLIP scores, thus there is room for learning and additional fine-tuning steps are beneficial. With $5.9\times$ more steps, the full-scale runs reach scores of 24.526 and 24.272, with the selection strategies achieving a BiomedCLIP score between $19.258$-$21.237$. On ROCO, EigDiff exhibits the largest gap (0.425) over the random strategy on PixArt-$\alpha$. Overall, EigDiff outperforms its competitors in six out of eight cases, with a selection that is $3.4$-$1607\times$ faster than the other selection strategies (see Table \ref{tab:selection_time}).   
\noindent
\textbf{Ablation studies. }
We conduct multiple ablation studies to evaluate the individual components of our method and their associated hyperparameters. Specifically, we analyze the impact of the number of quantile bins ($B$), the second-stage ratio ($\delta$), the PCA thresholds for both stages ($\tau_1$ and $\tau_2$), and the sample size of Stage 1 ($n_1$). All experiments are executed on the Pick-a-Pic dataset using SD1.5. To isolate the effect of each parameter, the remaining variables are held constant at their optimal values (see Table~\ref{tab:dataset_schedules}). We run each experiment three times, reporting the average PickScore along with its standard deviation.

\begin{table}[h]
\caption{Ablation of our sample selection strategy on the Pick-a-Pic dataset. This analysis demonstrates the relative contribution of each stage, as well as the effectiveness of the variance-based selection method introduced in Eq.~\eqref{eq:variance}.}
\vspace{-0.15cm}
\begin{center}
\centering
\setlength{\tabcolsep}{4pt}
\fontsize{8}{9}\selectfont{
\begin{tabular}{|c|c|c|c|c|}
\hline
\multirow{2}{*}{\textbf{Stage 1}} & \multirow{2}{*}{\textbf{Stage 2}} & \textbf{Variance-based} & \multirow{2}{*}{\textbf{PickScore}}\\
& & \textbf{selection} & \\
\hline
\color{Green}{\checkmark} & \color{Green}{\checkmark} & \color{Green}{\checkmark} & $\textbf{20.705}_{\pm0.023}$ \\
 \color{Green}{\checkmark}& \color{Green}{\checkmark}&  \color{Red}{\ding{55}} & $20.678_{ \pm 0.023}$ \\
 \color{Green}{\checkmark}&\color{Red}{\ding{55}}& \color{Green}{\checkmark} &  $20.703_{\pm0.022}$\\
  \color{Red}{\ding{55}}&\color{Green}{\checkmark}& \color{Green}{\checkmark} &  $20.685_{\pm 0.013}$\\
 \color{Red}{\ding{55}} & \color{Red}{\ding{55}} & \color{Red}{\ding{55}} & $20.669_{\pm0.013}$ \\
\hline
\end{tabular}
}
\end{center}
\label{tab:stagestrat}
\vspace{-0.1cm}
\end{table}

To isolate the contribution of each component, in Table~\ref{tab:stagestrat} we present an ablation study of our method. We evaluate configurations that use only Stage 1, only Stage 2, or a uniform instead of a variance-based sample distribution (Eq.~\eqref{eq:variance}), where $k_i =\Bigl\lfloor \frac{n_1}{|\mathcal{A}|}\Bigr\rfloor, \forall i\in \mathcal{A}$. The results demonstrate that Stage 1 provides the dominant preference signal, while Stage 2 introduces essential diversity for training stability. Ultimately, all partial configurations outperform random selection, while the complete method achieves the best results.

In Tables~\ref{tab:bins_abl} and~\ref{tab:ratio_abl}, we evaluate the impact of the quantile bin count ($B$ in Eq.~\eqref{eq:nobins}) and the inter-stage sample ratio ($\delta = \frac{n_2}{n_1}$). Modifying this ratio is functionally equivalent to ablating the diversity budget $n_2$. Empirically, we find that the $B=4$ and $\delta = 6$ yield the best overall results.

In Table \ref{tab:threshpcas}, we ablate the two thresholds $\tau_1, \tau_2$ from Eq. \eqref{eq:pc_quantiles}, indicating that in the case of Pick-a-Pic with SD1.5, the configuration of $\tau_1 = 0.6$ and $\tau_2 = 0.8$ is ideal. In Table \ref{tab:n1ablation}, we ablate the Stage 1 budget $n_1$ and, as expected, the score increases proportionally with the number of samples.

Finally, in Figure \ref{fig:placeholder}, we run both the eigen and random selection on 50,000 steps and compare with the full-scale baseline. The random approach almost continuously underperforms both the eigen and full-scale variant. Although EigDiff initially struggles against the full-scale baseline, in later steps, EigDiff becomes the clear winner. This explains why EigDiff and random selection perform subpar against the full-scale baseline on Pick-a-Pic with SD1.5 in Table \ref{tab:results_DPO}, as they do not benefit for a similar number steps to converge. With a matching number of epochs and less data, all data selection strategies use less overall training steps. While more optimization steps can increase the overall training time of our approach, EigDiff may outperform the full-scale baseline.

\begin{table}[t]
\caption{Ablation on the number of bins $B$ employed in Eq.~\eqref{eq:nobins}. Results are reported on the Pick-a-Pic dataset using SD1.5.}
\label{tab:bins_abl}
\vspace{-0.2cm}
\begin{center}
\centering
\setlength{\tabcolsep}{4pt}
\fontsize{8}{9}\selectfont{
\begin{tabular}{|c|c|}
\hline
$B$ & \textbf{PickScore} \\
\hline
1 & $20.675{\pm 0.003}$ \\
2 & $20.699\pm0.013$ \\
3 & $20.672\pm0.009$ \\
4 & $\textbf{20.705}\pm0.023$ \\
5 & $20.689\pm0.003$ \\
10 & $20.671 \pm 0.023$ \\
\hline
\end{tabular}
}
\end{center}
\vspace{-0.15cm}
\end{table}

\begin{table}[t]
\caption{Ablation on the sample ratio $\delta = \frac{n_2}{n_1}$.  Results are reported on the Pick-a-Pic dataset using SD1.5.}
\label{tab:ratio_abl}
\vspace{-0.2cm}
\begin{center}
\centering
\setlength{\tabcolsep}{4pt}
\fontsize{8}{9}\selectfont{
\begin{tabular}{|c|c|}
\hline
\textbf{Ratio} & \textbf{PickScore} \\
\hline
2 & $20.611\pm0.01$ \\
3 & $20.638\pm0.018$ \\
4 & $20.639\pm0.018$ \\
5 & $20.664\pm0.012$ \\
6 & $\textbf{20.705}\pm0.023$ \\
\hline
\end{tabular}
}
\end{center}
\vspace{-0.1cm}
\end{table}

\begin{table}[t]
\caption{Ablation on PCA thresholds for the two stages of selection.  Results are reported on the Pick-a-Pic dataset using SD1.5.}
\label{tab:threshpcas}
\centering
\vspace{-0.2cm}
\begin{center}
\setlength{\tabcolsep}{4pt}
\fontsize{8}{9}\selectfont{
\begin{tabular}{|c|c|}
\hline
\textbf{Thresholds} $(\tau_1,\tau_2)$ & \textbf{PickScore} \\
\hline
(0.2, 0.4) & $20.695\pm0.014$ \\
(0.4, 0.6) & $20.677\pm0.014$ \\
(0.6, 0.8) & $\textbf{20.705}\pm0.023$ \\
(0.8, 0.95) & $20.685\pm0.03$ \\
\hline
\end{tabular}
}
\end{center}
\vspace{-0.15cm}
\end{table}

\begin{table}[t]
\caption{Ablation on Stage 1 budget. Results are reported on the Pick-a-Pic dataset using SD1.5.}
\label{tab:n1ablation}
\centering
\vspace{-0.2cm}
\begin{center}
\setlength{\tabcolsep}{4pt}
\fontsize{8}{9}\selectfont{
\begin{tabular}{|c|c|}
\hline
$n_1$ & \textbf{PickScore} \\
\hline
512 & $20.627\pm0.013$ \\
1024 & $20.705\pm0.023$ \\
2048 & $20.777 \pm0.026$ \\
4096 & $\textbf{20.864}\pm0.013$ \\
\hline
\end{tabular}
}
\end{center}
\vspace{-0.15cm}
\end{table}

\vspace{-0.2cm}
\subsection{Qualitative Results}
\vspace{-0.2cm}

\begin{figure*}[!t]
    \centering
    \includegraphics[width=0.7\textwidth]{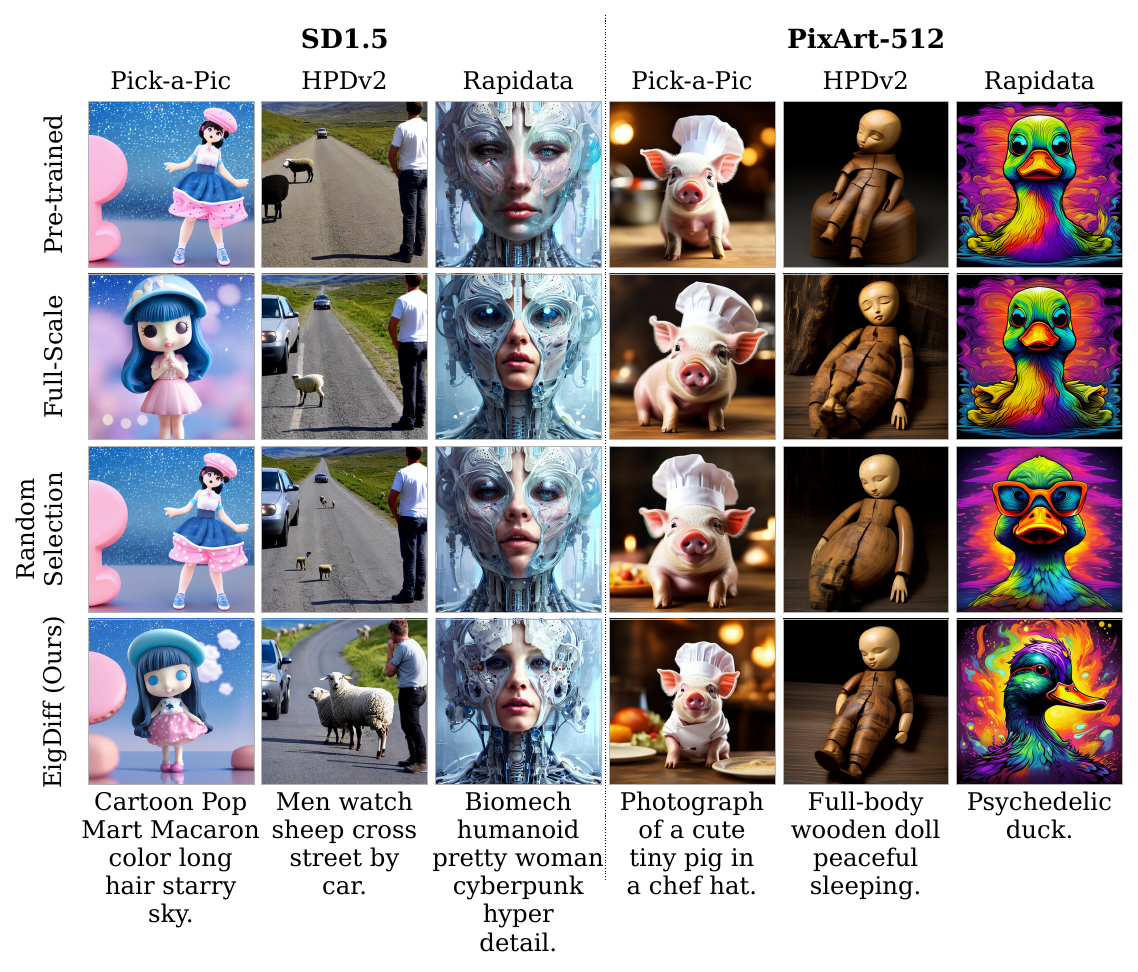}
    \vspace{-0.2cm}
    \caption{Qualitative results with the DPO regime across all datasets with SD1.5 and PixArt-$\alpha$. Best seen in color.}
    \label{fig:dpo_pdf}
\end{figure*}

\begin{figure*}[!t]
    \centering
    \includegraphics[width=0.9\textwidth]{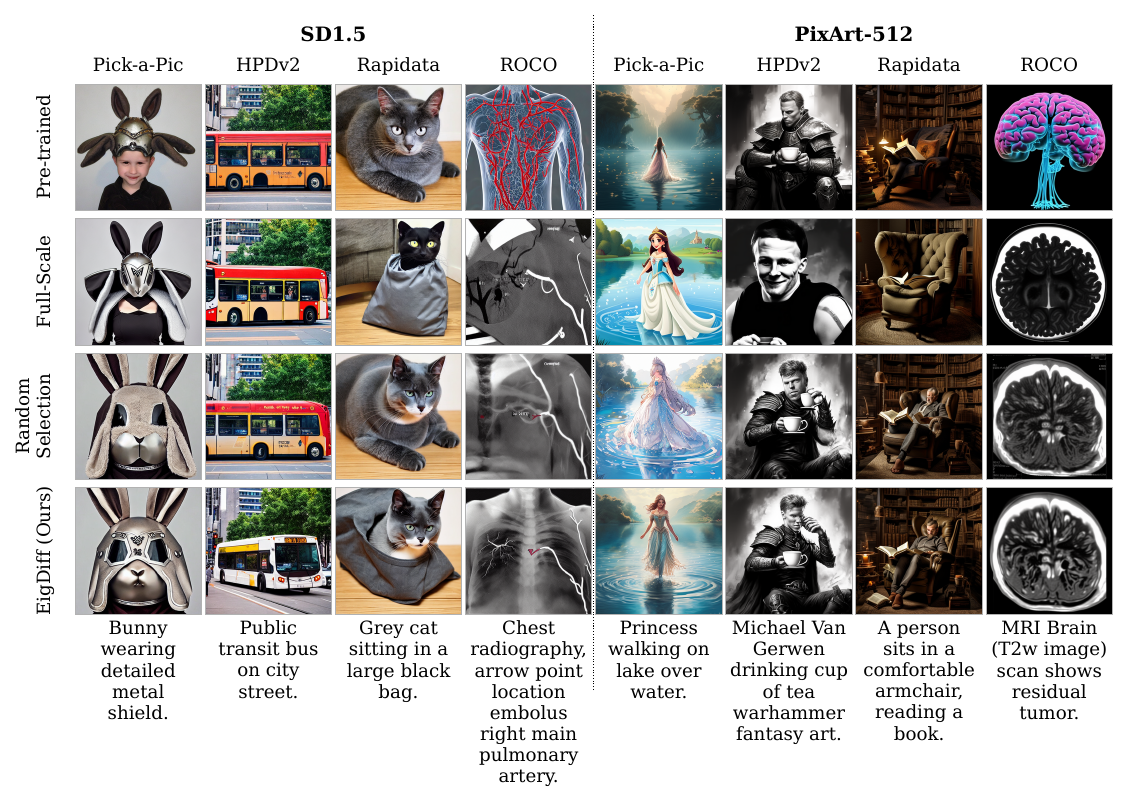}
    \vspace{-0.2cm}
    \caption{Qualitative results with the SFT regime across all datasets with SD1.5 and PixArt-$\alpha$. Best seen in color.}
    \label{fig:sft_pdf}
\end{figure*}

We present a qualitative analysis over DPO and SFT with all the baselines and our method in Figures \ref{fig:dpo_pdf} and \ref{fig:sft_pdf}, respectively. 

\noindent
\textbf{DPO visualization.} 
In the first column of Figure \ref{fig:dpo_pdf}, EigDiff aligns most closely with the prompt, whereas the full-scale model fails to generate the starry sky. Additionally, the random and pre-trained variants depict a Pop Mart figure with short hair, but fail to produce a high-quality macaroon. In the second column, EigDiff again achieves the best result, being the only strategy capable of generating a high-quality close-up of a sheep crossing the street with both a car and a man appropriately focused on the scene. The third and fourth columns similarly favor EigDiff, which produces the most prompt-aligned, high-quality images. Likewise, in the fifth column, EigDiff is the only method that avoids noticeable hallucinations. The pre-trained variant, for example, lacks arms and exhibits an awkward head shape and sleeping posture. Finally, in the last column, EigDiff produces the most recognizable psychedelic duck, with a well-defined side profile and vibrant, coherent coloration. In contrast, the pre-trained and full-scale variants fail to preserve the duck's structure, while the random selection generates a plausible frontal-view duck wearing glasses, but with lower overall definition and visual quality.   

\noindent
\textbf{SFT visualization.}
In the first column of Figure \ref{fig:sft_pdf}, EigDiff is the only method that accurately depicts a realistic bunny wearing a metal shield. In the second column, EigDiff is again the only strategy that generates a complete bus correctly framed, whereas the remaining methods hallucinate wheels or truncate the vehicle. The example shown on the third column favors EigDiff, which faithfully preserves both the gray cat and the large black bag specified in the prompt, while the competing methods either hallucinate the bag or swap the requested colors. In the fourth column, EigDiff produces the most accurate and detailed medical image, correctly depicting the radiograph, embolus annotation, and anatomical orientation, whereas the remaining variants exhibit substantial hallucinations. The example shown on the fifth column again favors EigDiff, as it is the only method that clearly conveys the princess walking. In the sixth column, despite the stronger pre-trained baseline, EigDiff clearly outperforms the full-scale and random variants by generating a realistic man holding a single cup of tea without major hallucinations. Similarly, in the seventh column, EigDiff is the only method that produces an anatomically coherent seated reader with minimal artifacts. Finally, the eighth column demonstrates that EigDiff generates the most detailed and faithful MRI depiction, while the other methods suffer from reduced quality and hallucinations, despite the full-scale model achieving higher quantitative scores on ROCO.

\vspace{-0.2cm}
\section{\uppercase{Conclusion}}
\vspace{-0.2cm}

In this paper, we presented EigDiff, a data-selection pipeline that embeds preference pairs with a frozen CLIP-based encoder, fits a PCA subspace on the preferred images, scores each pair by its signed alignment with the principal directions, and selects a compact subset via a two-stage, variance-proportional, stratified procedure.
We performed experiments with two diffusion backbones (SD1.5 and PixArt-$\alpha$), two fine-tuning regimes (DPO and SFT), and four datasets spanning natural and medical preference pairs with prompts. Our eigen selection method generally matches or exceeds the full-scale fine-tuning scores and surpasses all random experiments (at equal budget), while reducing GPU-minutes between $5.87\times$ and $13.95\times$ through subset selection. Against four selection baselines with an identical configuration, EigDiff is best or second-best in 12 of 14 cases, at a selection cost between $3.1\times$ and $1607\times$ lower. Our ablations show that the deterministic exemplars carry the dominant
preference signal, while the stratified secondary directions supply
non-redundant diversity. They also confirm that our variance-proportional budgeting has an important role in linking the first stage and the quantile-based second stage. The selection cost is a single PCA over cached embeddings, making the method a cheap, model-agnostic preprocessing step for preference-based diffusion fine-tuning. Limitations include reliance on the embedding geometry of the encoder. 

\vspace{-0.2cm}
\section{\uppercase{Acknowledgment}}
\vspace{-0.2cm}

This work was supported by a grant of the Ministry of Research, Innovation and Digitization, CNCS - UEFISCDI, project number PN-IV-P1-PCE-2023-0354, within PNCDI IV.

\bibliographystyle{apalike}
{\small
\bibliography{refs}}

\end{document}